# Higher Order Probabilities*
Henry E. Kyburg, Jr.

A number of writers have supposed that for the full specification of belief, higher order probabilities are required. Some have even supposed that there may be an unending sequence of higher order probabilities of probabilities of probabilities.... In the present paper we show that higher order probabilities can *always* be replaced by the marginal distributions of joint probability distributions. We consider both the case in which higher order probabilities are of the same sort as lower order probabilities and that in which higher order probabilities are distinct in character, as when lower order probabilities are construed as frequencies and higher order probabilities are construed as subjective degrees of belief. In neither case do higher order probabilities appear to offer any advantages, either conceptually or computationally.

1. Subjective probabilities are often introduced into systems of artifical intelligence because it is clear that some sort of uncertainty is required, and because it is unclear how else to represent that uncertainty. "Subjective" is used ambiguously. It may mean only that probabilities are to be relativized to subjects: that is, that any two rational (ideal) subjects having the same evidence will agree on probabilities. (Cheeseman [1985]) This corresponds to Keynes' notion of probability as a measure of rational belief (Keynes [1921]) Or "subjective" may be meant in a stronger sense: that there are no rules of rationality that can compel even <u>ideal</u> observers, having exactly the same information, to agree on probability. This was Savage's view, for example. (Savage [1954]) Many writers appear to have views more like Savage's than like Cheeseman's. This introduction of subjective probabilities in the strong sense, however, is quite often accompanied by a bad conscience: somehow we would like to have something better than mere subjective feeling to underlie our probabilities.

One way of easing one's conscience about the difference between assigning a probability to a head on a toss of a coin, and assigning a probability to a person's choice of a tie to go with a suit, is to consider second order probabilities. Loosely speaking, one says that the former probability is much more certain than the latter.

Savage himself admits to this feeling (pp. 57,58) and

30

characterizes it as a distinction between probabilities of which one "feels sure" and those of which one doesn't. He dismisses the feeling as useless, except as a guide to the <u>revision</u> of probabilities: When we find ourselves with degrees of belief that do not satisfy the probability calculus, we are moved to modify our degrees of belief; since there is no objectively correct way of proceeding to coherence, we do so in part by sacrificing probabilities about which we do not feel sure to probabilities about which we do feel sure.

The question of the meaningfulness of higher order probabilities has been discussed by a number of distinguished writers, including Savage, in Marshak *et al* [1975]. Chaim Gaifman [1985] and Zoltan Domotor [1981] both consider higher order probabilities as a way of extending probability to take account of uncertainties about probabilities. Richard Jeffrey, for whom probabilities are essentially derivable from preferences, considers higher order preferences (Jeffrey [1974]), from which one might think to get higher order probabilities. Brian Skyrms [1980a], [1980b] argues that higher order probabilities are essential for a correct representation of belief.

On the other hand, Cheeseman claims that "... information about the accuracy of $P$ is fully expressed by a probability density function over $P$." As an article of faith, this has a plausible ring to it. But the systems of Domotor and Gaifman, come with semantics that allow one to have actual models of systems with higher order probabilities. So higher order probabilities can certainly exist and be distinguished formally from first order probabilities. Brian Skyrms [1980a] and Hugh Mellor [1980] argue that in addition, higher order probabilities can reflect psychological realities that cannot be reflected by first order probabilities, and provide one way among others for characterizing the "laws of motion" of belief change, or probability kinematics.

For example, I might say that the probability that a coin will yield heads on a certain toss is "almost certainly" a half -- i.e., that the probability that the probability is a half is very close to one. In contrast, I might say that the probability that a certain person will choose a blue tie, given that she is wearing a blue suit, is 0.8, but I may be no

31

more than 50% confident of my probability judgement. That is, I might say that the probability that the probability is 0.8 is less than 0.5. The second order probabilities reflect my willingness to change my first order probabilities in the face of new evidence.

2.  In order to explore the question of whether higher order probabilities are useful for applications in AI, it will be helpful to approach these matters formally.

Let $W$ be our set of worlds, $w \in W$. Our initial or a *priori* probability function will be denoted by $P$. Disregarding considerations of higher order probabilities, our probability for a particular atom $w$ is $P(w)$ -- that represents the odds at which we would be willing to bet that $w$ was the case.

If we want to consider a second order probability, we must consider alternatives to our probability function $P$. ($P$ can't be wrong unless something else is right!) Let the second order probability function be denoted by $PP$. This is to be a classical probability function defined on a set $\mathcal{P}$ of classical probability functions whose common domain is $W$. There is an important relation between the first order probability $P$ and the second order probability $PP$. This has been noted by Jaynes [1958], Skyrms [1980], and others. The principle is that the first order probability $P(w)$ must be equal to the <u>expectation</u> of the second order probability applied to first order probabilities:

(1)    $P(w) = \sum_w PP(P_i) \times P_i(w) = E[P_i(w)]$

To see that this must so, reflect that the agent, were these two quantities not the same, would be rationally obligated to bet against himself for arbitrarily high stakes. Or, less picturesquely, that a cunning bettor could take advantage of him.

3.  There are two positions to take from which the question of higher order probabilities might get different answers. First, we might suppose that all probabilities are essentially the same -- for example, are expectation-forming

32

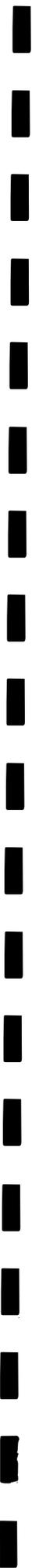

operators. Second, we might suppose that we distinguish two or more varieties of probability, and that "higher order" reflects an ordering among these varieties.

First, let us suppose that probability is univocal. If we construe probability univocally, then that probability must be the one we use for computing expectations, and, ultimately, for making decisions. Suppose we face a decision. The decision can be thought of as a choice from an exclusive and exhaustive set of acts $\mathcal{A}$. Associated with each act $A_j$ and each world $w$ is a utility $U(A_j, w)$. We suppose the sets of acts, worlds, and probability functions, to be finite, for the sake of simplicity.

If we knew the "correct" probability function $P^*$, the decision problem would be simple. We would just need to find an act $A_j$ such that no alternative act has a greater expected utility under $P^*$. $A_j$ is a correct decision just in case for all $k$,

(2) $\quad \sum_w P^*(w) \times U(A_j, w) \geq$

$\quad\quad\quad \sum_w P^*(w) \times U(A_k, w)$

Since we don't <u>know</u> what $P^*$ is, however, we must turn to second order probability. (We leave to one side here the intriguing question of what it means for a first order probability to be "correct".) $PP(P_i)$, which we may abbreviate $PP(i)$, is the second order probability that $P_i$ is the correct first order probability.

How does this change things? For one thing, it is clear that we get the same advice only if for every $w$, $P(w)$ is equal to the expected value of $P_i(w)$, as we observed in (1). In fact, this identity may be regarded as a constraint on second order probabilities. Our original equation (2), then, may be replaced by

(3) $\quad \sum_i PP(P_i) \times \left[ \sum_w P_i(w) \times U(A_j, w) \right] \geq$

$\quad\quad\quad \sum_i PP(P_i) \times \left[ \sum_w P_i(w) \times U(A_j, w) \right]$

This yields, by a trivial manipulation of the sums,

(4) $$\sum_{i,w} PP(P_i) \times P_i(w) \times U(A_j, w) \geq$$
$$\sum_{i,w} PP(P_i) \times P_i(w) \times U(A_k, w)$$

But in (4) it is apparent that what we have been calling 'first' and 'second' order probabilities are merely marginal probabilities of a distribution that we can represent as a probability distribution on $R = P \times W$ with probability element $P'(<i, w>) = PP(i) \times P(w)$ for $<i, w>$ in $P \times W$.

    Formally, this is no doubt the case. But is this just a formal trick? Can we make distinctive sense of the marginal probabilities that we are calling 'second order'? (Remember that we are not interpreting them in a different way as probabilities.) Since many joint probability distributions yield the same marginals, it is quite clear that there may be a loss of information in looking only at the marginal distributions. But we may also ask -- perhaps more importantly -- whether there is a computational advantage to this division of a joint probability distribution into the product of two marginal distributions.

    It turns out that we can express various useful things about the kinematics of certain marginal probabilities in terms of higher order probabilities. (This is reminiscent of the fact that in some special cases Dempster/Shafer conditionalization offers computational advantages over the convex Bayesian conditionalization of which it is a special case.) Here is an example taken from Skyrms [1980b].

    As is well known, Richard Jeffrey [1965] offers a procedure for updating a system of probabilities in response to a change in a given probability: If $P_i^*$ is an initial probability, $a$ a particular proposition, $P_f^*$ the final probability resulting from a shift exactly from $P_i^*(a)$ to $P_f^*(a)$ under the assumption that for all $b$ $P_i^*(b|a) = P_f^*(b|a)$, then for any $b$,

$$P_f^*(b) = P_i^*(b|a) \times P_f^*(a) + P_i^*(b|\sim a) \times P_f^*(\sim a)$$

This relation follows from certain constraints on higher order probabilities (Skyrms [1980b], appendix 2). The first two constraints essentially provide for the expected value condition we have already noted in (1); the third is this:

35

C3     $PP(b \mid a \,\&\, P(a) = x) = PP(b \mid a)$

This is a principle that seems appropriate for some contexts (where the conditional probability is based on known statistics) but inappropriate for others (where the object of our inquiry is that very conditional probability).

The upshot of this discussion is that if we construe first and second order probabilities in the same way, there is a perfectly automatic procedure for representing them as a joint distribution in a common space, but that it may involve a loss of information. There is no conceptual advantage to representing them as first and second order as opposed to joint. There _may_ be a computational advantage.

4.     Most people who have written about higher order probabilities have had in mind different _kinds_ of probabilities. Skyrms sometimes speaks of epistemic probabilities concerning relative frequencies or propensities, though he also talks of different orders of a given (epistemic) probability, as does D. H. Mellor [1980]. Domotor [1981] appears to consider a univocal notion of probability related to belief, but on close inspection the higher and lower order probabilities are not the same. Thus when we consider the probability that $A$ attributes to the probability that $B$ assigns to $A$'s having a certain probability for $a$, (Domotor's type of example), the probability functions are really all quite distinct.

To see how higher order probabilities work in this case, let us return to our original example. But let us make it more concrete: let us suppose that the worlds $w$ represent the different outcomes on the tenth toss of a die, and that the $P_i$ represent the various ways in which it may be loaded. Thus each $P_i$ is a sextuple of real numbers adding up to 1 that represent long-run relative frequencies or propensities, and $PP(P_i)$ is the degree of belief we have in the loading represented by the first order probability. (For simplicity, we suppose that we are certain that the outcomes of the tosses are independent and identically distributed.) This is about as clear a case as one can imagine in which the first and second order probabilities are of different kinds.

Suppose we have to choose between two actions: e.g., to bet at even money on the occurrence of a 'two' on the tenth roll, or to abstain from betting. The
computational procedure would be just that presented in
section 2, despite the fact that the probabilities appear to be
so different. We still can construct a product space, and a
joint distribution over it. Is this just an artifact? Are we
just mixing oil and water and calling it mayonnaise?

A careful look at the example shows that we are
not. What determines the utility of our act is not the
relative frequency of two's in general, but the relative
frequency of two's on the tenth roll -- i.e., whether there
is one or not. The $P_i$ 's give the long run frequency or the
propensity of the die to yield two's, but they do not in
general give the frequency of two's on the tenth toss.

There are many circumstances under which a
distribution such as that given by one of the $P_i$ would
determine the probability -- for example when we know
that the toss in question is an ordinary toss (not one
performed by someone who can control the outcome), that it
has not occurred yet, etc. The utility of an action under
the assumption of a particular loading hypothesis will, <u>under
these circumstances</u>, be determined by the the sextuple
embodied in that hypothesis. But this is just an instance of
what is traditionally called 'direct inference' from a statistical
distribution to a degree of belief. The conditions under which
direct inference is appropriate are just those under which it
is appropriate to weight the possible outcomes of the tenth
toss by the six numbers given by $P_i$ .

This is not the place to develop this argument (it has
been developed in various other places, e.g. [1974], [1985])
but we can summarize it as follows: knowing a statistical
distribution does not give us knowledge of the outcome of the
tenth toss; it just indicates (sometimes) how to allocate our
beliefs concerning the tenth toss. To choose among actions
whose outcomes depend on specific events requires beliefs; the
beliefs may depend on statistical knowledge. The second
order probabilities $PP(P_i)$ represent an allocation of our
beliefs among the possibilities indexed by $i$ . These may (or
may not) in turn be based on some form of statistical
knowledge, but the source of probabilities is irrelevant to the
question of whether it makes sense to combine them in a joint distribution. For a decision problem it clearly does make sense to combine them.

5. The conclusion is that so-called second order probabilities have nothing to contribute conceptually to the analysis and reprePsentation of uncertainty. The same ends can be achieved more simply, and without the introduction of novel machinery, by combining "first" and "second" order probabilities into a joint probability space. This procedure does not even add complexity to the computation, if the marginal distributions are independent. This is the case whether or not those probabilities are thought of as being of different kinds. Peter Cheeseman's claim that "information about the accuracy of $P$ is fully expressed by a probability density function over $P$," [1985, p 1007] appears to be fully vindicated, if construe "over $P$" to refer to a space in which all of our information can be expressed.

*This work has been supported in part by the Signals Warefare Center of the U. S. Army.